\documentclass{article}

    \usepackage[preprint]{neurips_2025}

\usepackage[utf8]{inputenc} 
\usepackage[T1]{fontenc}    
\usepackage{hyperref}       
\usepackage{url}            
\usepackage{booktabs}       
\usepackage{amsfonts}       
\usepackage{nicefrac}       
\usepackage{microtype}      
\usepackage{xcolor}         

\usepackage{graphicx}
\usepackage{multirow}
\usepackage{amsmath} 
\usepackage{threeparttable}
\usepackage{colortbl} 
\usepackage{graphicx}

\title{OmniIndoor3D: Comprehensive Indoor 3D Reconstruction}

\author{
    {\normalsize Xiaobao Wei$^{1}$\thanks{Equal contribution} \quad
    Xiaoan Zhang$^{1\ast}$ \quad
    Hao Wang$^{1}$ \quad
    Qingpo Wuwu$^{1}$} \\ 
    {\normalsize Ming Lu$^{1}$ \quad
    Wenzhao Zheng$^{2}$ \quad
    Shanghang Zhang$^{1}$\thanks{Corresponding author}}\\
    {\normalsize $^{1}$Peking University, Beijing, China \quad 
    $^{2}$University of California, Berkeley, USA}\\
    {\normalsize \texttt{weixiaobao0210@gmail.com}}
}

\begin{document}

\maketitle

\begin{abstract}
Indoor 3D reconstruction is crucial for the navigation of robots within indoor scenes. Current techniques for indoor 3D reconstruction, including truncated signed distance function (TSDF), neural radiance fields (NeRF), and 3D Gaussian Splatting (3DGS), have shown effectiveness in capturing the geometry and appearance of indoor scenes. However, these methods often overlook panoptic reconstruction, which hinders the creation of a comprehensive framework for indoor 3D reconstruction. To this end, we propose a novel framework for comprehensive indoor 3D reconstruction using Gaussian representations, called OmniIndoor3D. This framework enables accurate appearance, geometry, and panoptic reconstruction of diverse indoor scenes captured by a consumer-level RGB-D camera. Since 3DGS is primarily optimized for photorealistic rendering, it lacks the precise geometry critical for high-quality panoptic reconstruction. Therefore, OmniIndoor3D first combines multiple RGB-D images to create a coarse 3D reconstruction, which is then used to initialize the 3D Gaussians and guide the 3DGS training. To decouple the optimization conflict between appearance and geometry, we introduce a lightweight MLP that adjusts the geometric properties of 3D Gaussians. The introduced lightweight MLP serves as a low-pass filter for geometry reconstruction and significantly reduces noise in indoor scenes. To improve the distribution of Gaussian primitives, we propose a densification strategy guided by panoptic priors to encourage smoothness on planar surfaces. Through the joint optimization of appearance, geometry, and panoptic reconstruction, OmniIndoor3D provides comprehensive 3D indoor scene understanding, which facilitates accurate and robust robotic navigation. We perform thorough evaluations across multiple datasets, and OmniIndoor3D achieves state-of-the-art results in appearance, geometry, and panoptic reconstruction. We believe our work bridges a critical gap in indoor 3D reconstruction. The code will be released at: \href{https://ucwxb.github.io/OmniIndoor3D/}{https://ucwxb.github.io/OmniIndoor3D/}

\end{abstract}

\section{Introduction}
Robotic navigation is a crucial technique in embodied intelligence, necessitating a comprehensive reconstruction of the surrounding environment~\cite{alqobali2023survey, wijayathunga2023challenges}. The comprehensive reconstruction requires two essential components: spatial perception for obstacle avoidance and localization, and semantic understanding for manipulation and planning~\cite{crespo2020semantic, wu2024embodiedocc, wang2025embodiedoccpp}. Therefore, a comprehensive 3D reconstruction that captures appearance, geometry, and semantic features is essential for enabling indoor navigation.

Recent advances in 3D reconstruction, particularly Neural Radiance Fields (NeRF)~\cite{mildenhall2021nerf} and 3D Gaussian Splatting (3DGS)~\cite{kerbl20233d}, have led to significant breakthroughs in novel view synthesis and scene representation. Among these, 3DGS stands out because of its outstanding rendering quality and high efficiency~\cite{lee2024compact, lu2024scaffold, wei2024gazegaussian}. 
However, the lack of structured geometric constraints in 3DGS often leads to issues such as floating points, hallucinated surfaces, and incomplete reconstructions. 
Existing methods~\cite{turkulainen2025dn, guedon2024sugar, cheng2024gaussianpro, xiang2024gaussianroom, zhang20242dgs, yu2024gsdf} to address these limitations generally fall into two categories: (1) introducing explicit geometric regularization—e.g., Sugar~\cite{guedon2024sugar} enforces surface consistency through point-to-surface constraints; and (2) jointly optimizing an auxiliary model, such as a signed distance field (SDF), to guide Gaussian densification, as demonstrated in GaussianRoom~\cite{xiang2024gaussianroom}. Although these methods explore various strategies for achieving accurate appearance and geometric reconstruction, they overlook the panoptic reconstruction of the scene.

Panoptic scene understanding in 2D has achieved remarkable progress, driven by increasingly powerful network architectures and large-scale annotated datasets~\cite{kirillov2023segment, ravi2024sam, ren2024grounded}. 
However, extending panoptic segmentation from 2D to 3D remains a challenge. 
Unlike single-image segmentation, 3D panoptic segmentation~\cite{kirillov2019panoptic, xiong2019upsnet} requires consistent semantic and instance-level masks across multiple views, which demands a comprehensive understanding of the scene. 
NeRF-based methods~\cite{siddiqui2023panoptic, bhalgat2023contrastive, yu2024panopticrecon, yu2025leverage} utilize volumetric rendering to achieve promising panoptic segmentation but are limited by high computational cost. 
3DGS-based approaches~\cite{wu2024opengaussian, qin2024langsplat, ye2024gaussian, wang2024plgs, xie2025panopticsplatting} lift 2D segments or features into 3D space, enabling efficient open-vocabulary or panoptic segmentation. 
However, the precise boundaries for segmentation rely on a clear and accurate geometric reconstruction of the scene. These methods primarily focus on improving rasterized 2D segmentation quality while neglecting the regularization of 3D scene geometry, which ultimately limits their capability for comprehensive 3D scene reconstruction.  

To support robotic perception and planning, comprehensive 3D reconstruction is essential. Most existing methods address geometry and panoptic aspects separately, failing to recognize their interdependence. 
In addition, existing methods fail to exploit the mutual dependencies among appearance, geometry, and panoptic. The conflict between appearance and geometry reconstruction often hinders performance, while accurate geometry and panoptic reconstruction can reinforce each other. A comprehensive framework that simultaneously addresses all three aspects has not yet been thoroughly investigated.

\begin{figure*}[!t]
    \centering
    \includegraphics[width=1.0\textwidth]{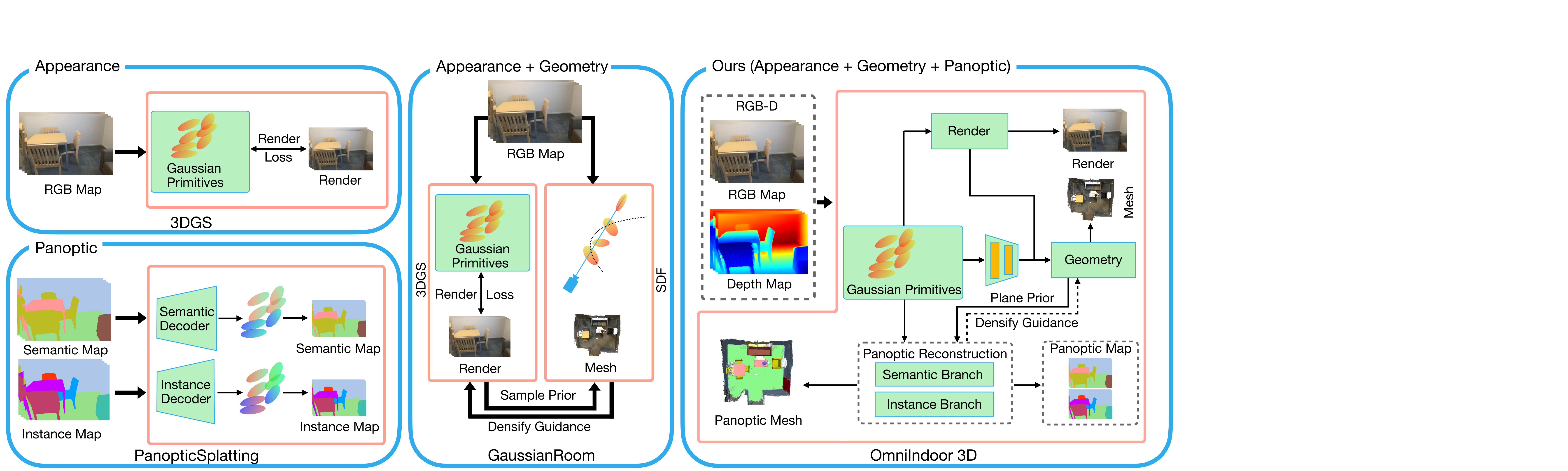}
    \vspace{-6mm}
    \caption{Comparison with existing methods. Unlike previous approaches that treat appearance, geometry, and panoptic reconstruction separately, our OmniIndoor3D presents a unified framework that leverages mutual dependencies for joint optimization, facilitating a comprehensive indoor 3D reconstruction essential for robotic navigation.}
    \label{fig:intro}
    \vspace{-4mm}
\end{figure*}

To bridge this gap, we propose OmniIndoor3D, the first framework to enable comprehensive indoor 3D reconstruction based on 3DGS. Given RGB-D images captured by a consumer-level camera, we first perform coarse reconstruction to initialize the Gaussian distribution of OmniIndoor3D. In vanilla 3DGS, jointly optimizing appearance and geometry leads to conflicts, primarily due to the entangled updates of Gaussian scales and rotations. Instead of simultaneously optimizing a time-consuming signed distance field (SDF), we introduce a lightweight multi-layer perceptron (MLP) that learns offsets for scale and rotation parameters. This MLP acts as a low-pass filter, suppressing high-frequency components in 3DGS and producing stable geometric properties, thereby decoupling geometry optimization from appearance refinement. To equip  OmniIndoor3D with panoptic reconstruction, we extend each Gaussian with semantic and instance features. These features are decoded into 3D panoptic labels via a semantic decoder and a set of 3D instance queries. 
Furthermore, to alleviate blur and noise in RGB-D observations, we propose a panoptic-guided densification strategy that adjusts the Gaussian gradients. The panoptic priors guide the spatial distribution of Gaussians, promoting both completeness and planar smoothness. 
Through end-to-end optimization, OmniIndoor3D jointly generates high-fidelity appearance, geometry, and panoptic reconstruction (Fig.~\ref{fig:intro}). Navigation robots equipped with standard RGB-D cameras can leverage OmniIndoor3D to achieve comprehensive indoor 3D understanding. 

Our contributions are summarized as follows: 1) We present OmniIndoor3D, a novel framework that achieves comprehensive indoor 3D reconstruction using Gaussian representation. 2) To decouple conflicts between geometry and appearance optimization, we propose a lightweight MLP to adjust the geometric properties of 3DGS. 3) To refine the Gaussian distribution and enhance planar smoothness, we introduce a panoptic-guided densification strategy to assist reconstruction with panoptic information. 4) Extensive experiments on ScanNet and ScanNet++ demonstrate that OmniIndoor3D achieves state-of-the-art performance in novel view synthesis, geometric reconstruction, and panoptic lifting.

\section{Related Work}
\noindent\textbf{Neural Scene Representation.} Neural scene representation can be categorized into NeRF-based and 3DGS-based methods. 
Neural Radiance Fields (NeRF)~\cite{mildenhall2021nerf} model scenes as continuous volumetric fields and have shown impressive results in novel view synthesis. 
Subsequent works improve rendering efficiency~\cite{muller2022instant, liu2020neural} and geometric fidelity by incorporating depth supervision, smoothness regularization, and multi-view consistency~\cite{deng2022depth, niemeyer2022regnerf, wang2023sparsenerf}. 
However, surfaces extracted from NeRF using Marching Cubes~\cite{lorensen1998marching} often remain noisy or incomplete. To address this, alternative representations such as occupancy grids~\cite{niemeyer2020differentiable} and TSDFs~\cite{wang2021neus, li2023neuralangelo} have been explored, often guided by SfM priors or geometric constraints~\cite{fu2022geo, yu2022monosdf}. 
Recently, 3D Gaussian Splatting (3DGS)~\cite{kerbl20233d} has emerged as a fast and expressive representation using learnable Gaussian primitives. While efficient for view synthesis, vanilla 3DGS typically relies on SfM initialization~\cite{schonberger2016structure}, leading to suboptimal geometry. 
To improve spatial distribution, some works utilize LiDAR or RGB-D point clouds~\cite{huang2024S3G, chen2024omnire}, while others introduce monocular priors such as depth and normals~\cite{xiang2024gaussianroom, zhang20242dgs}. 
Representation extensions include flattened or planar Gaussians~\cite{guedon2024sugar, huang20242d} and hybrid 2D-3D models~\cite{chen2024mixedgaussianavatar}. 
Another direction introduces SDF-constrained optimization~\cite{yu2024gsdf}, jointly refining Gaussians and SDF fields. 
CarGS~\cite{shen2025evolving} further decouples appearance and geometry by identifying scale and rotation as key conflict factors, proposing a geometry-aware MLP. 
Despite these advancements, most methods remain focused on appearance and geometry reconstruction, lacking the panoptic reconstruction that is critical for robotic perception tasks.


\noindent\textbf{Panoptic Segmentation and Lifting. }
Panoptic segmentation, first introduced in~\cite{kirillov2019panoptic}, aims to provide a unified understanding of object instances ("things") and semantic regions ("stuff") in diverse scenes. Early works like UPSNet~\cite{xiong2019upsnet} integrate panoptic segmentation into single networks with novel architectures, extending this concept to 3D has been critical for applications in autonomous driving and robotics. 
The evolution of 3D panoptic reconstruction has primarily followed two representation paradigms. Neural Radiance Field (NeRF)~\cite{mildenhall2021nerf} methods encode scenes into neural networks, offering implicit representations of 3D scene properties including appearance, geometry, and semantics. 
Several approaches employ NeRFs for 3D panoptic segmentation. 
Panoptic Lifting~\cite{siddiqui2023panoptic} proposes a label lifting scheme with linear assignment between predictions and unaligned instance labels. 
Contrastive Lift~\cite{bhalgat2023contrastive} achieves 3D object segmentation through contrastive learning. 
PVLFF~\cite{chen2024panoptic} builds an instance feature field for open-vocabulary segmentation. 
PanopticRecon~\cite{yu2024panopticrecon, yu2025leverage} introduces methods for aligning 2D masks and guiding 3D instance segmentation. 
Despite their promising performance, NeRF-based methods are computationally intensive and thus unsuitable for deployment in real-time robotics applications.
Alternatively, 3D Gaussian Splatting (3DGS)~\cite{kerbl20233d} methods optimize differentiable Gaussians with real-time rendering speed. 
Approaches like LEGaussians~\cite{shi2024language}, LangSplat~\cite{qin2024langsplat}, and Feature 3DGS~\cite{zhou2024feature} extend Gaussian splatting by adding feature attributes, while Gaussian Grouping~\cite{ye2024gaussian}, OpenGaussian~\cite{wu2024opengaussian}, and PLGS~\cite{wang2024plgs} focus on instance segmentation through various techniques. PanopticSplatting~\cite{xie2025panopticsplatting} proposes an end-to-end system for open-vocabulary panoptic reconstruction with query-guided instance segmentation. However, these semantic lifting or feature lifting methods are mainly designed to enhance 2D segmentation, while neglecting the geometric optimization required for accurate 3D semantic mesh reconstruction. 

We propose OmniIndoor3D, a unified framework that jointly conducts appearance, geometry, and panoptic reconstruction within an efficient Gaussian representation. By decoupling the conflict between appearance and geometry optimization, and leveraging panoptic cues to guide panoptic reconstruction, our method enables high-quality and meaningful indoor 3D reconstruction.


\begin{figure*}[!t]
\vspace{-10pt}
    \centering
    \includegraphics[width=1.0\textwidth]{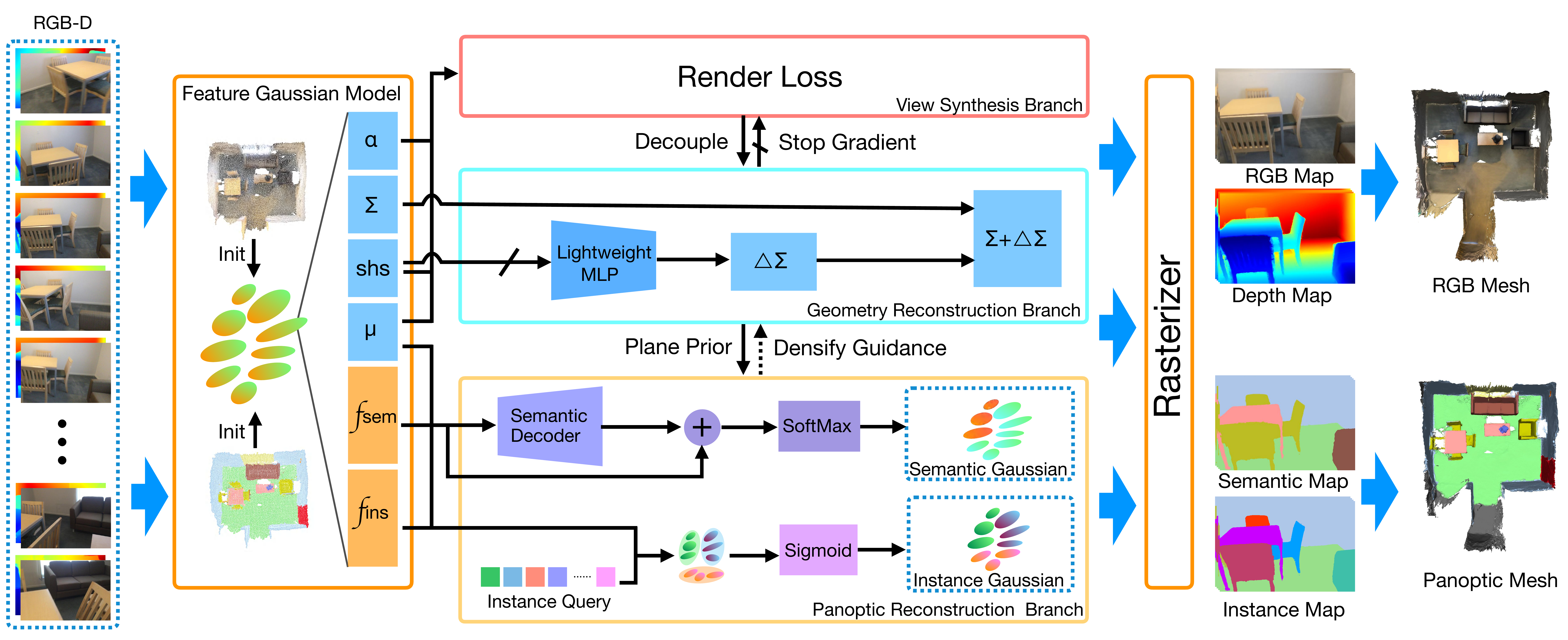}
    \vspace{-6mm}
    \caption{Pipeline of OmniIndoor3D. Given posed RGB-D as inputs, we first extract a coarse 3D reconstruction to initialize the Gaussian distribution. The network subsequently optimizes three dedicated branches, each responsible for novel view synthesis, geometric reconstruction, and panoptic lifting.}
    \label{fig:pipeline}
    \vspace{-6mm}
\end{figure*}

\vspace{-2mm}
\section{Method}
\vspace{-4mm}
\subsection{3D Gaussian Initialization and Representation} 
\vspace{-2mm}
Gaussian initialization is essential for stable optimization and high-quality rendering. While vanilla 3DGS relies on sparse SfM point clouds, COLMAP is often slow and inaccurate, especially in complex indoor environments. Instead, we utilize RGB-D images captured by consumer-level sensors to reconstruct a coarse but structured point cloud. Multi-view depth maps are projected into the world coordinate frame and aggregated to form a unified point cloud. A voxelization step is applied to reduce noise and control point density. The resulting point cloud is then used to initialize the spatial position $\mu$ and the appearance-related spherical harmonics coefficients $shs$ of the Gaussian primitives. 
To initialize the semantic feature $f_{{sem}}$ and instance feature $f_{{ins}}$ of each Gaussian, we follow PanopticSplatting~\cite{xie2025panopticsplatting} by using Grounded SAM~\cite{ren2024grounded} to extract pseudo semantic and instance labels from the input images. These labels are projected into 3D space using the corresponding depth maps, resulting in a 3D point cloud with coarse semantic and instance annotations. We use these labeled points to assign initial semantic and instance features to the Gaussians.

After initialization, the indoor scene is represented as a set of 3D Gaussian primitives. Each Gaussian is defined as a tuple
$G = (\Sigma, \mu, shs, \alpha, f_{{sem}}, f_{{ins}})$,
where $\mu \in \mathbb{R}^3$ is the center, $\Sigma \in \mathbb{R}^{3\times3}$ is the covariance matrix, $shs \in \mathbb{R}^{3(k+1)^2}$ denotes the view-dependent color represented by spherical harmonics of degree $k=3$, and $\alpha \in \mathbb{R}$ is the opacity. In addition, each Gaussian carries a semantic feature $f_{{sem}} \in \mathbb{R}^{N_{{sem}}}$ and an instance feature $f_{{ins}} \in \mathbb{R}^{N_{{ins}}}$, where $N_{{sem}}$ is the number of semantic classes, and $N_{{ins}}$ is set to a value larger than the maximum number of instances present in the scene. 

The spatial density of a Gaussian centered at $\mu$ is defined as:
\begin{equation}
    G(x)=e^{-\frac{1}{2}(x-\mu)^T\Sigma^{-1}(x-\mu)},
\end{equation}
where $x$ is a 3D position in the scene. To ensure that the covariance matrix $\Sigma$ is positive semi-definite, it is constructed as $\Sigma = R S S^\top R^\top$, where $S \in \mathbb{R}^3$ is a diagonal scaling matrix and $R \in \mathbb{R}^{3 \times 3}$ is a rotation matrix.
Each 3D Gaussian is associated with an opacity value $\alpha$, which modulates its spatial contribution $G(x)$ during the blending process. This weighted contribution is used in both rendering and reconstruction tasks. 3DGS enables efficient scene rendering through tile-based rasterization, avoiding traditional ray marching. Each 3D Gaussian $G(x)$ is projected onto the image plane as a 2D Gaussian, and a tile-based rasterizer composites the scene via $\alpha$-blending: 
\begin{equation}
    F(x^{\prime})=\sum_{i\in N}f_i\sigma_i\prod_{j=1}^{i-1}(1-\sigma_j),\quad\sigma_i=\alpha_iG_i(x),
\end{equation}
where $x'$ denotes a pixel location in the image plane, and $N$ is the set of 2D Gaussians overlapping that pixel after projection. The value $f_i$ depends on the target task: it is set to the view-dependent color $shs_i$ for appearance rendering, to the depth $z_i$ for geometry reconstruction, and to the semantic or instance feature $f_{{sem}}$ or $f_{{ins}}$ for panoptic segmentation. 

\subsection{Appearance and Geometry Reconstruction}
\vspace{-2mm}
Inspired by prior works~\cite{shen2025evolving} identifying the Gaussian covariance as a primary source of conflict between appearance and geometry optimization, we propose to decouple geometry reconstruction via a dedicated lightweight MLP. Specifically, we introduce a geometry-specific covariance adjustment module that acts as a low-pass filter to suppress high-frequency noise in indoor scenes. This module is implemented as a single MLP, which takes as input the detached appearance features and view direction, and predicts a 7D vector representing the scale and rotation (as a quaternion) used to construct the covariance matrix. To avoid mutual interference between the appearance and geometry branches, we denote the input appearance feature as a detached vector $\phi$, extracted from the spherical harmonics coefficients $shs$. Formally, the geometry-adjusted covariance is computed as:
\begin{equation}
    \Delta \Sigma = M^{geo}_\Sigma(\phi, \theta),
\end{equation}
where $\phi$ is the detached appearance feature, $\theta$ is the view direction, and $M^{geo}_\Sigma$ is the geometry MLP. The output $\Delta \Sigma \in \mathbb{R}^7$ consists of three scale parameters and four rotation parameters, which are used to construct the geometry-specific covariance matrix for each Gaussian. This formulation enables the geometry branch to refine structural consistency without affecting rendering quality, and proves effective in reducing geometric noise while maintaining surface smoothness. 
After optimization, we render the RGB image from the optimized Gaussians.
The depth map is rendered using the covariance adjusted by the geometry MLP, and is then used to construct a TSDF volume for mesh extraction. 
\vspace{-2mm}
\subsection{Panoptic Reconstruction}
\vspace{-2mm}
\noindent\textbf{Semantic Branch. } Inspired by prior works~\cite{xie2025panopticsplatting} on 3D-aware semantic segmentation, we introduce a semantic decoding module that maps each Gaussian's semantic feature $f_{{sem}}$ into a class probability. To better leverage the pseudo labels assigned during initialization, we design a residual prediction strategy. Specifically, a semantic MLP $M_{{sem}} $takes as input semantic features $f_{{sem}}$ and position $\mu$ of the Gaussians, and predicts a correction term to refine the initial semantic logits. Formally, the predicted class logits for one Gaussian are computed as:
\begin{equation}
    l_{{sem}} = \text{softmax}(f_{{sem}} + M_{{sem}}(f_{{sem}}, \mu)),
\end{equation}
where $l_{{sem}} \in \mathbb{R}^{N_c}$ is the normalized class probability vector for $N_c$ semantic categories. This residual formulation reduces optimization difficulty and stabilizes learning by preserving the initialization prior. Once the semantic labels of Gaussians are obtained, we perform label blending during the rasterization process. Instead of blending raw features as in~\cite{wu2024opengaussian}, we directly blend the predicted class probabilities using the $\alpha$-blending rule: 
\begin{equation}
    M_{sem}(x') = \sum_{i \in \mathcal{N}} l_{{sem}}^{(i)} \, \alpha'_i \prod_{j=1}^{i-1} (1 - \alpha'_j),
\end{equation}
where $x'$ is a pixel on the image plane, $\mathcal{N}$ is the set of visible Gaussians projected onto $x'$, and $\alpha'_i$ denotes the normalized opacity of the $i$-th Gaussian. This label blending strategy emphasizes 3D consistency by classifying Gaussians in 3D space before projecting to 2D, which helps mitigate the influence of noisy 2D supervision. Additionally, the softmax normalization prevents Gaussians within the camera frustum from overwhelming the prediction. 

\noindent\textbf{Instance Branch.} We adopt a query-based design for instance segmentation, inspired by ~\cite{xie2025panopticsplatting}, where a set of learnable instance queries interact with scene features to cluster Gaussians into instances. Each instance query is composed of a feature vector $f_q \in \mathbb{R}^{d}$ and a 3D position $p_q \in \mathbb{R}^3$. The covariance matrix $\Sigma_q \in \mathbb{R}^{3 \times 3}$ of the instance query is constructed from its scale and rotation parameters. This enables a geometry-aware attention mechanism, where the affinity between a scene Gaussian $g$ and an instance query $q$ is computed by jointly considering feature similarity and spatial proximity. 

The attention weight between a scene Gaussian and a query is defined as:
\begin{equation}
    A(q, g) = \text{sim}(f_q, f_{{ins}}^{(g)}) \cdot \mathcal{G}(p_g; p_q, \Sigma_q),
\end{equation}
where $\text{sim}(f_q, f_{{ins}}^{(g)})$ denotes feature similarity (e.g., dot product), and $\mathcal{G}(p_g; p_q, \Sigma_q)$ is the probability density of the query’s 3D Gaussian evaluated at the scene Gaussian center $p_g$. $\Sigma_q$ is constructed from the query’s scale and rotation parameters. We apply a softmax over all queries to obtain the instance label distribution for each Gaussian:
\begin{equation}
    l_{{ins}}(g) = \text{softmax}\left( \{ A(q_i, g) \}_{i=1}^N \right),
\end{equation}
where $N$ is the total number of instance queries. The final 2D instance map is then rendered via $\alpha$-blending of these Gaussian-level labels:
\begin{equation}
    M_{ins}(x') = \sum_{i \in \mathcal{N}} l_{{ins}}^{(i)} \alpha_i' \prod_{j=1}^{i-1} (1 - \alpha_j'),
\end{equation}
where $\mathcal{N}$ denotes the sorted list of Gaussians projected to pixel $x'$.
To improve computational efficiency, we adopt local cross-attention by restricting the query-Gaussian interaction to Gaussians located within the view frustum. Notably, both the semantic and instance rendering processes utilize the geometry-adjusted covariance. The geometry branch provides accurate planar priors that enhance the quality and consistency of segmentation. 
\vspace{-2mm}
\subsection{Panoptic-guided Densification}
\vspace{-2mm}
We further improve the spatial distribution of Gaussians through a panoptic-guided densification strategy. Previous geometry-based methods~\cite{chen2024pgsr} primarily rely on signed distance fields (SDF) to control Gaussian growth. However, these approaches neglect the rich semantic priors available in the scene. To address this, we introduce semantic priors into the densification process. Specifically, we compute a confidence-aware SDF modulation by weighting each Gaussian’s SDF value with its semantic confidence. Let $s = d(x) - z(x)$ denote the signed distance function (SDF) value at Gaussian center $x$,  
where $d(x)$ is the rendered depth sampled from the image plane, and $z(x)$ is the projected depth of the Gaussian center along the camera ray. 
The semantic confidence is obtained by taking the softmax over the semantic logits $f_{{sem}}$ and selecting the maximum non-background class probability. The modulation is defined as:
\begin{equation}
\zeta(s) = \exp\left( -\frac{s^2}{2\sigma^2} \right), \quad \epsilon_g = \nabla_g + \omega_g \cdot \zeta(s) \cdot c_{{sem}},
\end{equation}
where $\nabla_g$ is the accumulated gradient magnitude of a Gaussian, $c_{{sem}}$ is the semantic confidence, and $\omega_g$ controls the influence of geometric guidance. A new Gaussian is spawned when $\epsilon_g$ exceeds a fixed threshold. This design encourages densification in semantically meaningful and geometrically uncertain regions, improving coverage and segmentation completeness in challenging indoor scenes. 
\vspace{-2mm}
\subsection{Training}
\vspace{-2mm}
\noindent\textbf{Appearance Loss.} We supervise the rendered image $I$ using a combination of L1 and SSIM losses with respect to the ground truth image $I^{gt}$:
\begin{equation}
    \mathcal{L}_{rgb} = (1 - \lambda_{SSIM})\mathcal{L}_1(I, I^{gt}) + \lambda_{SSIM}\mathcal{L}_{SSIM}(I, I^{gt}),
\end{equation}
\noindent\textbf{Geometry Loss.} To enforce geometric consistency, we supervise the rendered depth $D(x)$ using the ground-truth depth $D^{gt}(x)$ captured by the RGB-D camera:
\begin{equation}
    \mathcal{L}_{{depth}} = \frac{1}{|\mathcal{W}|} \sum_{x \in \mathcal{W}} \|D(x) - D^{gt}(x)\|_1,
\end{equation}
To further improve global consistency, we introduce a cross-view loss. A pixel $x_r$ in the reference view is projected to a neighboring view via homography $H_{rn}$ and then back-projected using $H_{nr}$. The consistency is enforced by minimizing the forward-backward reprojection error:
\begin{equation}
    \mathcal{L}_{cross} = \frac{1}{|\mathcal{W}|} \sum_{x_r \in \mathcal{W}} \|x_r - H_{nr}H_{rn}x_r\|,
\end{equation}
\noindent\textbf{Panoptic Loss.} For semantic supervision, a cross-entropy loss is applied between the rendered semantic logits $M_{sem}$ and ground truth $M_{sem}^{gt}$:
\begin{equation}
    \mathcal{L}_{sem} = \mathcal{L}_{ce}(M_{sem}, M_{sem}^{gt}),
\end{equation}
For the instance branch, we adopt a combination of Dice loss and binary cross-entropy (BCE) loss between the predicted instance map $M_{ins}$ and the ground truth $M^{gt}_{ins}$:
\begin{equation}
    \mathcal{L}_{ins} = \mathcal{L}_{dice}(M_{ins}, M_{ins}^{gt}) + \mathcal{L}_{bce}(M_{ins}, M_{ins}^{gt}),
\end{equation}
\noindent\textbf{Total Loss.} The total training objective is a weighted sum of all loss components:
\begin{equation}
    \mathcal{L}_{total} = \lambda_{{rgb}} \mathcal{L}_{rgb} + \lambda_{{depth}} \mathcal{L}_{depth} + \lambda_{{cross}} \mathcal{L}_{cross} + \lambda_{{sem}} \mathcal{L}_{sem} + \lambda_{{ins}} \mathcal{L}_{ins},
\end{equation}
In our experiments, the weights are empirically set as $\lambda_{{rgb}}=1.0$, $\lambda_{{depth}}=1.0$, $\lambda_{{cross}}=1.5$, $\lambda_{{sem}}=0.5$, and $\lambda_{{ins}}=0.5$.

\section{Experiments}
\subsection{Experimental Setup}
\textbf{Datasets.} We validate our approach on two publicly available indoor scene datasets, ScanNet~\cite{dai2017scannet} and ScanNet++~\cite{yeshwanth2023scannet++}. 
To assess reconstruction and rendering quality, and compare with current state-of-the-art methods, we select the same scenes as GaussianRoom~\cite{xiang2024gaussianroom}, a total of 10 indoor scenes, 8 scenes from ScanNet, and 2 scenes from ScanNet++. To evaluate the panoptic segmentation performance, we follow PanopticSplatting~\cite{xie2025panopticsplatting} and select 7 indoor scenes, including 4 from ScanNet and 3 from ScanNet++. We strictly follow the experimental settings used in the baseline methods, including the training and validation splits as well as the evaluation tools. More details on dataset preprocessing are provided in the appendix.

\textbf{Evaluation metrics.} We adopt standard evaluation metrics for each task. For novel view synthesis, we report SSIM, PSNR, and LPIPS to measure image quality. For geometric reconstruction, we use Accuracy (Acc.), Completion (Com.), Precision (Pre.), Recall (Re.), and F1-score (F1). For panoptic lifting, we evaluate with PQ, SQ, RQ, mIoU, mAcc, mCov, and mW-Cov. Due to space limitations, detailed definitions of all metrics are provided in the appendix.

\begin{table}[!t]
\caption{Quantitative comparison for novel view synthesis. Results are averaged over the same selected scenes as in GaussianRoom.}
\label{tab:2D_compare_quality}
\centering
\scalebox{0.9}
{\begin{tabular}{lccc|ccc}
\toprule
\multirow{2}{*}{\textbf{Method}} & \multicolumn{3}{c}{ScanNet} & \multicolumn{3}{c}{ScanNet++} \\ 
\cmidrule(r){2-4} \cmidrule(r){5-7}
 & SSIM$\uparrow$ & PSNR$\uparrow$ & LPIPS$\downarrow$ & SSIM$\uparrow$ & PSNR$\uparrow$ & LPIPS$\downarrow$ \\ 
\midrule
3DGS~\cite{kerbl20233d} & 0.731  & 22.133 & 0.387 & 0.843 & 21.816 & 0.294 \\
SuGaR~\cite{guedon2024sugar} & 0.737 & 22.290 & 0.382 & 0.831 & 20.611 & 0.318 \\
GaussianPro~\cite{cheng2024gaussianpro} & 0.721 & 22.676 & 0.395 & 0.831 & 21.285 & 0.320 \\
DN-Splatter~\cite{turkulainen2025dn} & 0.639 & 21.621 & 0.312 & 0.826 & 20.445 & 0.268 \\
GaussianRoom~\cite{xiang2024gaussianroom} & 0.758 & 23.601 & 0.391 & 0.844 & 22.001 & 0.296 \\ 
\midrule
\textbf{Ours} & \textbf{0.812} & \textbf{25.817} & \textbf{0.304} & \textbf{0.879} & \textbf{25.139}  & \textbf{0.193} \\ \bottomrule
\end{tabular}
}
\vspace{-0.1cm}
\end{table}
\begin{figure}[!t]
  \centering
  \includegraphics[width=1.0\linewidth]{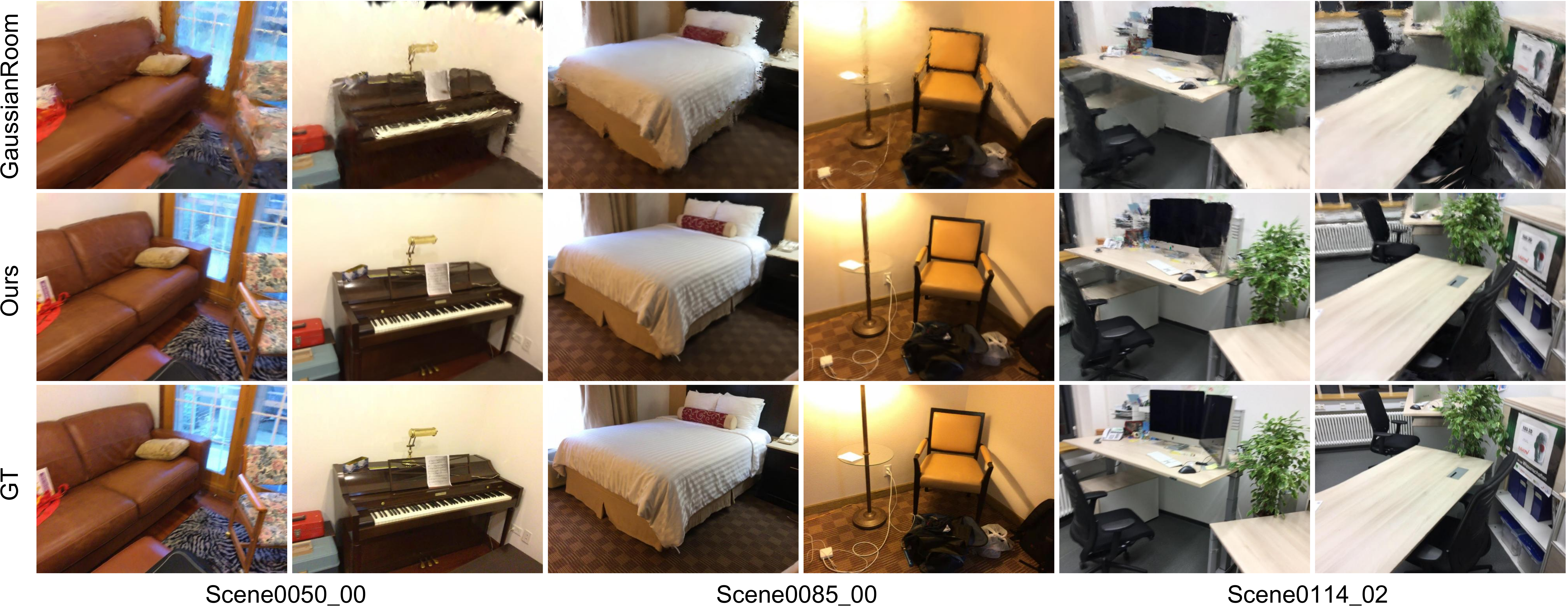}
\vspace{-6mm}
\caption{Visualization comparison for novel view synthesis. }
\vspace{-4mm}
  \label{fig:novel_view_synthesis}
\end{figure}

\subsection{Novel View Synthesis}

We evaluate OmniIndoor3D on novel view synthesis against leading 3DGS-based surface reconstruction methods. As shown in Tab.~\ref{tab:2D_compare_quality}, our approach consistently outperforms baselines across metrics on both ScanNet and ScanNet++. GaussianRoom~\cite{xiang2024gaussianroom} suffers from low performance, as its SDF-guided pruning removes detail-preserving Gaussians and causes over-smoothing. 
Our improved performance stems from two key components: (1) RGB-D fusion for Gaussian initialization provides structured geometric priors, reducing noise and ambiguity during early optimization; (2) a lightweight MLP, guided by depth regularization, adjusts Gaussian geometry and effectively decouples appearance from geometry optimization. This prevents high-frequency noise and preserves visual fidelity. As shown in Fig.~\ref{fig:novel_view_synthesis}, OmniIndoor3D generates sharper textures and more consistent multi-view structures, while GaussianRoom often suffers from rendering holes due to over-pruning.

\begin{table}[!t]
\caption{Quantitative comparison for geometric reconstruction. Results are averaged over the same
selected scenes as in GaussianRoom.}
\vspace{-2mm}
\label{tab:recon}
\centering
\scalebox{0.8}
{
\begin{tabular}{lccccc|ccccc}
\toprule
\multirow{2}{*}{\textbf{Method}} & \multicolumn{5}{c}{ScanNet} & \multicolumn{5}{c}{ScanNet++} \\
\cmidrule(r){2-6} \cmidrule(r){7-11}
 & Acc.$\downarrow$ & Com.$\downarrow$ & Pre.$\uparrow$ & Re.$\uparrow$ & F1$\uparrow$ & Acc.$\downarrow$ & Com.$\downarrow$ & Pre.$\uparrow$ & Re.$\uparrow$ & F1$\uparrow$ \\ 
\midrule
COLMAP~\cite{schonberger2016structure} & 0.062 & 0.090 & 0.640 & 0.569 & 0.600 & 0.091 & 0.093 & 0.519 & 0.520 & 0.517 \\ 
\midrule
NeRF~\cite{mildenhall2021nerf} & 0.160 & 0.065 & 0.378 & 0.576 & 0.454 & 0.135 & 0.082 & 0.421 & 0.569 & 0.484 \\
NeuS~\cite{wang2021neus} & 0.105 & 0.124 & 0.448 & 0.378 & 0.409 & 0.163 & 0.196 & 0.316 & 0.265 & 0.288 \\
MonoSDF~\cite{yu2022monosdf} & 0.048 & 0.068 & 0.673 & 0.558 & 0.609 & 0.039 & 0.043 & 0.816 & 0.840 & 0.827 \\
HelixSurf~\cite{liang2023helixsurf} & 0.063 & 0.134 & 0.657 & 0.504 & 0.567 & ------ & ------ & ------ & ------ & ------ \\ 
\midrule
3DGS~\cite{kerbl20233d} & 0.338 & 0.406 & 0.129 & 0.067 & 0.085 & 0.113 & 0.790 & 0.445 & 0.103 & 0.163 \\
GaussianPro~\cite{cheng2024gaussianpro} & 0.313 & 0.394 & 0.112 & 0.075 & 0.088 & 0.141 & 1.283 & 0.353 & 0.081 & 0.129 \\
SuGaR~\cite{guedon2024sugar} & 0.167 & 0.148 & 0.361 & 0.373 & 0.366 & 0.129 & 0.121 & 0.435 & 0.444 & 0.439 \\ 
DN-Splatter~\cite{turkulainen2025dn} & 0.212 & 0.210 & 0.153 & 0.182 & 0.166 & 0.294 & 0.276 & 0.108 & 0.108 & 0.107 \\ 
2DGS~\cite{huang20242d} & 0.167 & 0.152 & 0.311 & 0.341 & 0.324 & ------ & ------ & ------ & ------ & ------ \\
GaussianRoom~\cite{xiang2024gaussianroom} & 0.047 & 0.043 & 0.800 & 0.739 & 0.768 & 0.035 & 0.037 & 0.894 & 0.852 & 0.872 \\ 
\midrule
\textbf{Ours} & \textbf{0.023} & \textbf{0.024} & \textbf{0.927} & \textbf{0.907} & \textbf{0.917} & \textbf{0.008} & \textbf{0.011} & \textbf{0.996} & \textbf{0.970} & \textbf{0.983} \\ 
\bottomrule
\end{tabular}
}
\vspace{-2mm}
\end{table}
\begin{figure}[!t]
  \centering
  \includegraphics[width=1.0\linewidth]
                                   {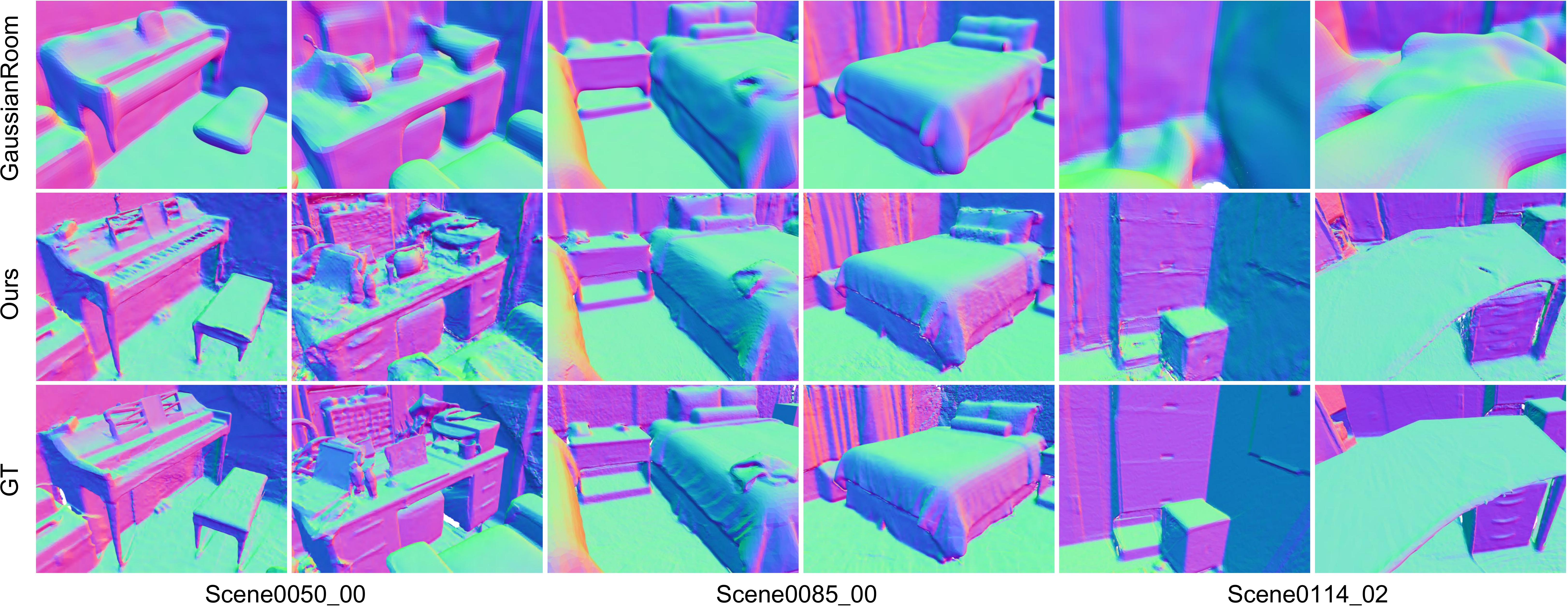}
\vspace{-6mm}
  \caption{Visualization comparison for geometric reconstruction. }
\vspace{-6mm}
  \label{fig:rencon}
\end{figure}

\subsection{Geometric Reconstruction}
\vspace{-2mm}
As shown in Tab.~\ref{tab:recon}, our method outperforms all baselines and achieves state-of-the-art results in all evaluated scenes. 
OmniIndoor3D significantly improves metrics across both datasets, indicating that our approach preserves more accurate surface details while maintaining structural completeness. These improvements can be attributed to the use of RGB-D fusion for Gaussian initialization and our depth-guided geometric regularization. Additionally, we adopt a panoptic-guided densification strategy to densify Gaussians near planar regions. This encourages Gaussians to be cloned and split at appropriate locations. 
From the visual comparisons in Fig.~\ref{fig:rencon}, we observe that our method reconstructs clearer object boundaries and finer surface details. Mesh reconstructed by OmniIndoor3D exhibits fewer noisy fluctuations and better preservation of flat and planar structures. In contrast, GaussianRoom suffers from over-smoothed surfaces and distorted geometry, particularly in thin or high-frequency regions. This degradation stems from its reliance on an SDF field for mesh extraction. 
Moreover, our framework remains efficient. Instead of relying on an additional SDF field with high memory and computational cost, we employ a lightweight MLP that reduces the interference of geometry optimization on appearance rendering. 

\begin{table*}[!t]
\caption{Quantitative comparison for panoptic lifting. Results are averaged over the same
selected scenes as in PanopticSplatting.}
\centering
\setlength{\tabcolsep}{5pt}
\renewcommand\arraystretch{1.2}
\resizebox{\linewidth}{!}{
\begin{tabular}{lccc|cc|cc|ccc|cc|cc}
\toprule
\multirow{2}{*}{\textbf{Method}} & \multicolumn{7}{c}{ScanNet} & \multicolumn{7}{c}{ScanNet++} \\
\cmidrule(r){2-8} \cmidrule(r){9-15}
 & PQ$\uparrow$ & SQ$\uparrow$ & RQ$\uparrow$ & mIoU$\uparrow$ & mAcc$\uparrow$ & mCov$\uparrow$ & mW-Cov$\uparrow$ & PQ$\uparrow$ & SQ$\uparrow$ & RQ$\uparrow$ & mIoU$\uparrow$ & mAcc$\uparrow$ & mCov$\uparrow$ & mW-Cov$\uparrow$ \\
\midrule
Panoptic Lifting~\cite{siddiqui2023panoptic}
    & 57.86 & 61.96 & 85.31 & 67.91 & 78.59 & 45.88 & 59.93 
    & 71.14 & 77.48 & 88.14 & 81.34 & 89.67 & 56.17 & 68.51 \\
Contrastive Lift~\cite{bhalgat2023contrastive}
    & 37.35 & 41.91 & 57.60 & 64.77 & 75.80 & 13.21 & 23.26
    & 47.58 & 57.23 & 65.81 & 81.09 & 89.30 & 27.39 & 36.51 \\
PVLFF~\cite{chen2024panoptic}
    & 30.11 & 51.71 & 44.43 & 55.41 & 63.96 & 45.75 & 48.41
    & 52.24 & 66.86 & 65.56 & 62.53 & 70.31 & 67.95 & 75.47 \\
PanopticRecon~\cite{yu2024panopticrecon}
    & 63.70 & 64.81 & 81.17 & 68.62 & 80.87 & 66.58 & 77.84
    & 68.29 & 77.01 & 85.05 & 77.75 & 87.08 & 51.34 & 62.79 \\
\midrule
Gaussian Grouping~\cite{ye2024gaussian}
    & 43.75 & 50.63 & 72.68 & 58.05 & 68.68 & 52.70 & 58.10
    & 33.10 & 40.60 & 67.27 & 59.53 & 68.13 & 29.83 & 36.83\\
OpenGaussian~\cite{wu2024opengaussian}
    & 48.73 & 51.48 & 88.10 & 54.05 & 68.43 & 44.43 & 49.60 
    & 51.03 & 56.93 & 85.73 & 61.80 & 73.97 & 50.00 & 51.02  \\
PanopticSplatting~\cite{xie2025panopticsplatting}
    & 74.75 & 74.75 & \textbf{100.0} & 74.95 & 83.70 & 73.18 & 79.63
    & 77.73 & 82.70 & 93.60 & 81.90 & 89.50 & 74.73 & 78.03 \\
\midrule
\textbf{Ours}
    & \textbf{84.60} 
    & \textbf{88.74}
    & 94.69
    & \textbf{76.19}
    & \textbf{87.42}
    & \textbf{81.08}
    & \textbf{83.96}
    & \textbf{86.12}
    & \textbf{89.51}
    & \textbf{94.85}
    & \textbf{89.27}
    & \textbf{90.23}
    & \textbf{76.33}
    & \textbf{79.82}\\

\bottomrule
\end{tabular}
}

\label{tab:seg_scannet}
\vspace{-8mm}

\label{tab:Comparative_baseline}
\end{table*}

\begin{figure}[!t]
  \centering
  \includegraphics[width=1.0\textwidth]
                                   {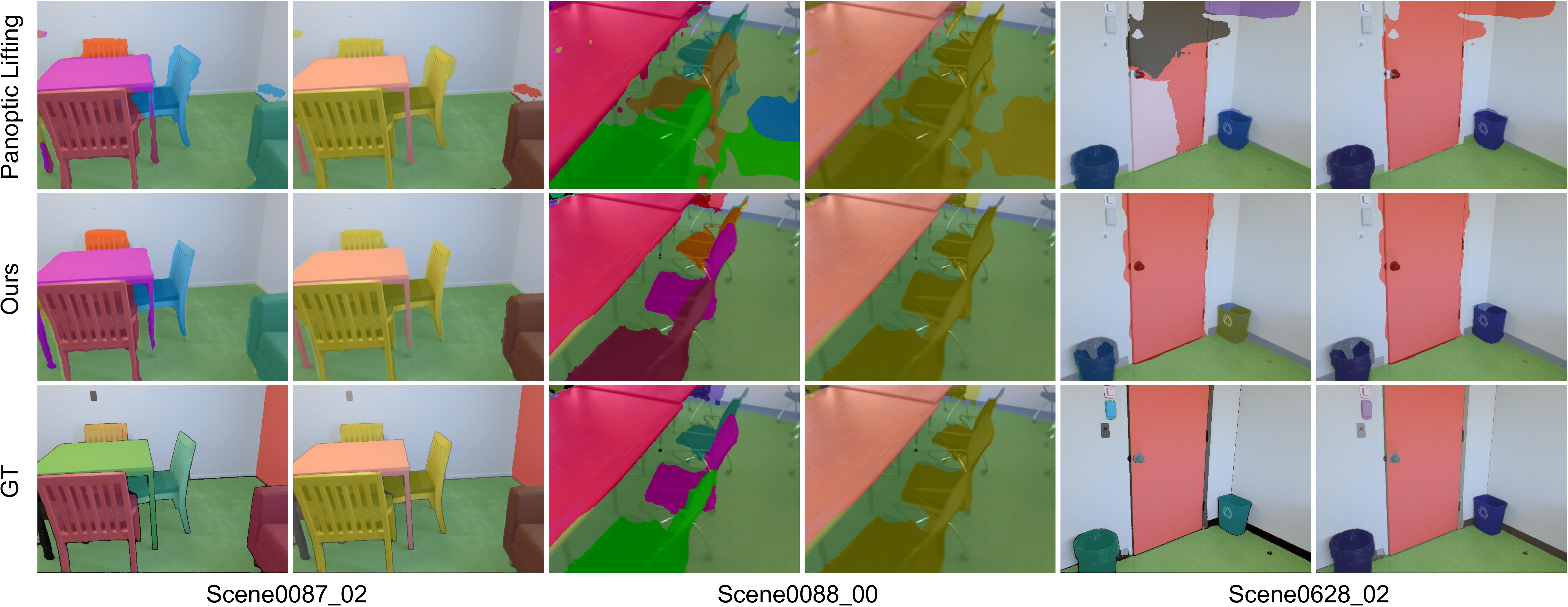}
\vspace{-6mm}
  \caption{Visualization comparison for panoptic lifting.}
\vspace{-6mm}
  \label{fig:semantic}
\end{figure}

\vspace{-2mm}
\subsection{Panoptic Lifting}
\vspace{-2mm}
We evaluate the panoptic lifting performance of our method on the ScanNet and ScanNet++ datasets. As shown in Tab.~\ref{tab:seg_scannet}, our approach outperforms all baselines across different evaluation metrics. 
Compared to Panoptic Lifting~\cite{siddiqui2023panoptic} and Contrastive Lift~\cite{bhalgat2023contrastive}, our method achieves significantly better segmentation quality. These NeRF-based methods suffer from slow training and rendering time. 3DGS-based methods such as OpenGaussian~\cite{wu2024opengaussian} and PanopticSplatting~\cite{xie2025panopticsplatting} improve efficiency but still struggle with accurate geometry. These models mainly lift 2D predictions into 3D, leading to inconsistent instance boundaries and fragmented labels. 
As illustrated in Fig.~\ref{fig:semantic}, our method delivers sharper instance boundaries and more complete segmentations. This improvement comes from our RGB-D fusion for Gaussian initialization and panoptic-guided densification strategy. By incorporating semantic priors during optimization, we regularize the Gaussian distribution and promote planar consistency. 
Overall, OmniIndoor3D provides a unified solution for appearance, geometry, and panoptic reconstruction. 

\begin{table*}[!ht]
\vspace{-4mm}
\caption{Component-wise ablation study. Evaluated on the same scenes as in Tab.~\ref{tab:seg_scannet}.}
\label{tab:ablation}
\centering
\setlength{\tabcolsep}{3pt}
\renewcommand\arraystretch{1.2}
\resizebox{\linewidth}{!}{
\begin{tabular}{lccc|ccccc|ccccccc}
\toprule
\multirow{2}{*}{\textbf{Method}} & \multicolumn{3}{c}{\textbf{Novel View Synthesis}} & \multicolumn{5}{c}{\textbf{Geometric Reconstruction}} & \multicolumn{7}{c}{\textbf{Panoptic Segmentation}} \\
\cmidrule(r){2-4} \cmidrule(r){5-9} \cmidrule(r){10-16}
 & SSIM$\uparrow$ & PSNR$\uparrow$ & LPIPS$\downarrow$ & Acc.$\downarrow$ & Com.$\downarrow$ & Pre.$\uparrow$ & Re.$\uparrow$ & F1$\uparrow$ & PQ$\uparrow$ & SQ$\uparrow$ & RQ$\uparrow$ & mIoU$\uparrow$ & mAcc$\uparrow$ & mCov$\uparrow$ & mW-Cov$\uparrow$ \\
\midrule
w/o RGB-D Init & 0.844 & 25.728 & 0.315 & 0.102 & 0.039 & 0.581 & 0.796 & 0.663 & 37.25 & 45.27 & 52.77 & 68.95 & 72.48 & 50.31 & 58.05 \\
w/o Geo Decouple & 0.874 & 29.899 & 0.263 & 0.021 & 0.019 & 0.936 & 0.938 & 0.937 & 74.68 & 88.55 & 80.84 & 77.72 & 84.77 & 68.46 & 77.92 \\
w/o Pan-guided & 0.858 & 30.356 & 0.285 & 0.024 & \textbf{0.017} & 0.934 & \textbf{0.949} & 0.941 & 80.55 & 87.74 & 87.74 & \textbf{77.80} & 78.82 & 67.49 & 76.60 \\
\midrule
\textbf{Full Model} & \textbf{0.890} & \textbf{30.956} & \textbf{0.250} & \textbf{0.019} & 0.018 & \textbf{0.949} & 0.948 & \textbf{0.948} & \textbf{84.60}  & \textbf{88.74} & \textbf{94.69} & 76.19 & \textbf{87.42} & \textbf{81.08} & \textbf{83.96} \\

\bottomrule
\end{tabular}
}
\vspace{-4mm}
\end{table*}
\vspace{-2mm}
\subsection{Component-wise Ablation Study}
\vspace{-2mm}

To assess the contribution of each key component in OmniIndoor3D, we conduct an ablation study across the three core tasks: novel view synthesis, geometric reconstruction, and panoptic lifting on the 4 ScanNet scenes. The results are summarized in Tab.~\ref{tab:ablation}. 
Removing RGB-D initialization leads to a disorganized Gaussian distribution and reduced reconstruction quality, as the lack of structured depth priors introduces ambiguity in early optimization. 

Without the proposed MLP-based geometry decoupling module, appearance and geometry optimization interfere with each other, causing degradation across all tasks. Although geometry remains relatively accurate, the rendering performance decreases. Panoptic segmentation also suffers from the conflict between them, highlighting the importance of separating the two objectives.

Disabling panoptic-guided densification primarily affects panoptic segmentation by reducing semantic coverage and consistency, showing that semantic priors are essential for guiding Gaussian growth toward meaningful regions. 
Additional ablation studies are provided in the appendix.

\vspace{-4mm}
\section{Conclusion}
\vspace{-2mm}
We introduce OmniIndoor3D, the first unified framework for comprehensive indoor 3D reconstruction, which simultaneously performs appearance, geometry, and panoptic reconstruction using Gaussian representations.
Extensive experiments across multiple benchmarks demonstrate that OmniIndoor3D achieves state-of-the-art performance in novel view synthesis, geometric reconstruction, and panoptic lifting. 
Our approach enables robust and consistent 3D scene understanding, facilitating improved robotic navigation within complex indoor environments.



\newpage

\bibliographystyle{plain}
\bibliography{ref.bib}

\end{document}